\documentclass[sn-mathphys-num]{sn-jnl}

\usepackage{graphicx}
\usepackage{multirow}
\usepackage{amsmath,amssymb,amsfonts}
\usepackage{amsthm}
\usepackage{mathrsfs}
\usepackage[title]{appendix}
\usepackage{xcolor}
\usepackage{textcomp}
\usepackage{manyfoot}
\usepackage{booktabs}
\usepackage{algorithm}
\usepackage{algorithmicx}
\usepackage{algpseudocode}
\usepackage{listings}

\usepackage[most]{tcolorbox}

\usepackage{color}
\usepackage{soul}
\definecolor{lightyellow}{rgb}{0.97, 0.97, 0}
\sethlcolor{lightyellow}

\usepackage{placeins}

\begin{document}

	\title{Ontology-based Semantic Similarity Measures for Clustering Medical Concepts in Drug Safety}
	
	
	\author[1]{\fnm{Jeffery L.} \sur{Painter}}\email{jeffery.l.painter@gsk.com}
	\author[2]{\fnm{François} \sur{Haguinet}}\email{francois.f.haguinet@gsk.com}
	\author[1]{\fnm{Greg} \sur{Powell}}\email{gregory.e.powell@gsk.com}
	\author[3,4]{\fnm{Andrew} \sur{Bate}}\email{andrew.x.bate@gsk.com}

	\affil*[1]{\orgname{GlaxoSmithKline}, \orgaddress{\city{Durham}, \state{NC}, \country{USA}}}
	\affil[2]{\orgname{GlaxoSmithKline}, \orgaddress{\city{Wavre}, \country{Belgium}}}
	\affil[3]{\orgname{GlaxoSmithKline}, \orgaddress{\city{London}, \country{UK}}}
	\affil[4]{\orgname{London School of Hygiene and Tropical Medicine}, \orgaddress{\city{London}, \country{UK}}}

	
	\abstract{Semantic similarity measures (SSMs) are widely used in biomedical research but remain underutilized in pharmacovigilance. This study evaluates six ontology-based SSMs for clustering MedDRA Preferred Terms (PTs) in drug safety data. Using the Unified Medical Language System (UMLS), we assess each method’s ability to group PTs around medically meaningful centroids. A high-throughput framework was developed with a Java API and Python/R interfaces support large-scale similarity computations. Results show that while path-based methods perform moderately with F1 scores of 0.36 for WUPALMER and 0.28 for LCH, intrinsic information content (IC)-based measures, especially INTRINSIC\_LIN and SOKAL, consistently yield better clustering accuracy (F1 Score of 0.403). Validated against expert review and standard MedDRA queries (SMQs), our findings highlight the promise of IC-based SSMs in enhancing pharmacovigilance workflows by improving early signal detection and reducing manual review.}

	
	\maketitle

	\section{Introduction}
		
		Semantic similarity measures (SSMs) are essential tools in biomedical informatics, enabling applications in genomics, ontology enrichment, and phenotypic clustering \cite{lord2003investigating}. They support functional clustering of genes and proteins \cite{pesquita2009semantic, guzzi2012semantic} and have been used to classify diseases from structured clinical data, including electronic health records (EHRs) \cite{martinez2013semantic, hauben2009defining}. Despite their success in genomics and clinical phenotyping, SSMs remain underutilized in pharmacovigilance (PV). This is surprising, as the ability to cluster similar cases effectively is critical for identifying safety issues with medicines as early as possible \cite{bate2012terminological}. Moreover, no single level of specificity is optimal across all use cases in the hierarchical terminologies commonly used in PV \cite{bate2012terminological}. The ability to use semantic distance to cluster safety events, therefore, has considerable potential to improve medicine and vaccine safety.
		
		PV relies heavily on standardized terminologies— WHODrug\footnote{\url{https://who-umc.org/whodrug/whodrug-global/}}  for medicinal products and MedDRA for adverse events (AEs) \cite{fescharek2004medical}. MedDRA plays a central role in aggregating clinically similar outcomes across patients using a rigid four-level hierarchy. However, this structure often fails to reflect more nuanced or mechanistic relationships between terms \cite{brown2003methods}. For example, \textit{neutropenia} (10029354) and \textit{bone marrow failure} (10065553)—both indicators of myelosuppression—reside in entirely separate High-Level Group Term (HLGT) branches. Widely adopted safety data mining practices typically analyze data at the preferred term (PT) level, which would lead to these two terms being treated independently \cite{bate2012terminological}.
		
		In addition, MedDRA imposes a binary grouping model: AEs are either in the same category or not, with no graded scale of similarity and no mechanism to incorporate evolving clinical knowledge or terminology changes. This lack of granularity limits MedDRA’s ability to capture the broader semantic landscape that underlies many safety concepts.
		
		Semantic similarity offers a potential solution by quantifying the degree of relatedness between terms on a continuous scale. Rather than relying solely on fixed hierarchies, semantic distance measures—particularly when derived from integrated ontologies—can identify previously unlinked but clinically relevant AEs. Dupuch and Grabar \cite{dupuch2015semantic} demonstrated this by mapping MedDRA PTs into SNOMED-CT \cite{snomed} hierarchies via UMLS\footnote{\url{https://www.nlm.nih.gov/research/umls/index.html}}, creating the \textit{ontoEIM} resource. This enriched semantic network introduced intermediate nodes and decomposed complex AEs into semantic primitives (e.g., gastric ulcer → ulcer + stomach), yielding more fine-grained clusters than MedDRA alone. Their study also showed improved alignment with Standardized MedDRA Queries (SMQs) \cite{mozzicato2007standardised} and enhanced signal detection performance.
		
		Recent work has further demonstrated the value of data-driven semantic representations for PV. Gattepaille \cite{gattepaille2019using} and Erlanson et al. \cite{erlanson2025clinical} applied distributional embedding models (vigiVec) to learn vector representations of AEs and drugs from spontaneous reports in WHO’s VigiBase. These embeddings captured clinically coherent relationships—often outperforming MedDRA and ATC groupings in tasks such as intruder detection, an evaluation task in which the model must identify the least related term in a group. However, these models derive similarity from co-occurrence patterns in observational data and therefore reflect empirical rather than ontological proximity. These results provide early evidence that semantic approaches—whether ontology-based or corpus-derived—can outperform traditional hierarchical groupings in clustering and relatedness tasks.
		
		In contrast, our approach computes semantic distance from formal biomedical ontologies, leveraging the hierarchical structures and curated relationships encoded in UMLS. We integrated MedDRA, SNOMED-CT, and MeSH \cite{dhammi2014medical} to construct comprehensive semantic networks and computed both path-based and information content (IC)-based similarity metrics using an optimized implementation of Apache cTAKES YTEX \cite{savova2010mayo}. Unlike corpus-based models, this ontology-grounded framework assesses concept relatedness based on expert-defined medical knowledge structures rather than usage patterns. To support reproducible, large-scale evaluations, we developed a RESTful API with Python and R interfaces, enabling scalable similarity computations across extensive concept sets.
		
		We benchmarked the performance of multiple SSMs by clustering MedDRA Preferred Terms (PTs) around clinically meaningful centroids. Although we initially explored algorithmic centroid detection (e.g., selecting the PT with the highest average similarity to others in an SMQ), we adopted a more interpretable approach: using narrow SMQs that contain a PT matching the SMQ’s name as centroids. This enabled consistent evaluation and straightforward comparison across methods. Cluster validity was assessed against both SMQs and expert medical review. Our evaluation demonstrates that semantic similarity measures (SSMs), when carefully applied to drug safety data, can facilitate the clustering of Preferred Terms (PTs) around medically relevant centroids, offering a scalable, ontology-driven alternative to traditional hierarchical groupings.		

	\section{Methods}
	
		\subsection{Semantic Similarity Measures Evaluated}
		
			To identify effective semantic similarity measures (SSMs) for clustering safety-related concepts, we explored ontological representations of MedDRA and SNOMED-CT, integrated via UMLS. From this, we selected six ontology-based SSMs leveraging hierarchical structure and information content (IC): two path-based measures—Leacock and Chodorow (LCH) \cite{fellbaum1998combining} and Wu and Palmer (WUPALMER) \cite{wu1994verb}—and four IC-based measures—Resnik (INTRINSIC\_RESNIK) \cite{resnik1995using}, Lin (INTRINSIC\_LIN) \cite{lin1998information}, an intrinsic variant of LCH (INTRINSIC\_LCH), and Sokal (SOKAL) \cite{sokal1962comparison, sanchez2011semantic}. Each measure’s formal definition follows standard implementations and is outlined below. 
				
			\subsubsection{Leacock and Chodorow (LCH)}
			The Leacock and Chodorow measure computes similarity based on the shortest path length between two concepts in the taxonomy, scaled by the maximum depth of the hierarchy. Due to the logarithmic transformation, it is not constrained between 0 and 1:
			\clearpage
			\[
			\text{LCH}(c_1, c_2) = -\log\left(\frac{\text{length}(c_1, c_2)}{2 \times \text{max\_depth}}\right).
			\]
			
			\subsubsection{Wu and Palmer (WUPALMER)}
			The Wu and Palmer measure quantifies similarity by considering the depths of the two concepts and their lowest common ancestor (LCA) in the hierarchy:
			\[
			\text{WUPALMER}(c_1, c_2) = \frac{2 \times \text{depth}(\text{LCA}(c_1,c_2))}{\text{depth}(c_1) + \text{depth}(c_2)}.
			\]
			
			\subsubsection{Resnik (INTRINSIC\_RESNIK)}
			Resnik’s measure is based on the IC of the LCA of two concepts. IC is typically derived from concept occurrence frequencies in a corpus or computed intrinsically from ontology structure. Like LCH, this metric is unbounded:
			\[
			\text{RESNIK}(c_1,c_2) = IC(\text{LCA}(c_1,c_2)).
			\]
			
			\subsubsection{Lin (INTRINSIC\_LIN)}
			Lin’s measure extends Resnik’s approach by incorporating the IC of the individual concepts, yielding a normalized similarity score:
			\[
			\text{LIN}(c_1,c_2) = \frac{2 \times IC(\text{LCA}(c_1,c_2))}{IC(c_1) + IC(c_2)}.
			\]
			
			\subsubsection{Intrinsic Leacock and Chodorow (INTRINSIC\_LCH)}
			This variant modifies the LCH measure by incorporating intrinsic IC, adjusting path-based similarity using IC-weighted scaling. Although not widely standardized, we refer to it as INTRINSIC\_LCH for clarity. Like LCH, it is not constrained between 0 and 1.
			\[
			\text{INTRINSIC}_{\text{LCH}(c_1, c_2)} = -\log \left( \frac{IC(c_1) + IC(c_2) - 2 \times IC(\text{LCA}(c_1, c_2)) + 1}{2 \times \max IC} \right)
			\]

			\subsubsection{Sokal (SOKAL)}
			The Sokal similarity measure applies IC-based transformations like Lin but incorporates a specific normalization factor. It can be interpreted as an IC-based distance measure reformulated into a similarity metric \cite{sanchez2011semantic}.
			\[
			\text{SOKAL}(c_1, c_2) = 
			\frac{IC\left(\text{LCA}(c_1, c_2)\right)}
			{2 \times \left(IC(c_1) + IC(c_2)\right) - 3 \times IC\left(\text{LCA}(c_1, c_2)\right)}
			\]
			
		\subsection{Evaluation of SSMs}
		
			We assessed SSMs by their ability to cluster Preferred Terms (PTs) around centroid PTs. Centroids were drawn from Standardized MedDRA Queries (SMQs) and internal safety-related PT groupings from GSK product global data sheets. Due to limited utility of higher MedDRA hierarchy levels, we focused on narrow SMQs (20–50 PTs) containing a PT identical to the SMQ name, declaring this PT the centroid for clarity. Fifteen SMQs met this criterion, of which 12 overlapped with internal GSK-defined safety concept sets.
		
			For each of these 12 centroid PTs, we computed pairwise similarity scores using the six selected SSMs across all PTs in MedDRA v26.1.

		\subsection{Metric Selection}
		
			The feasibility study initially assessed 10 semantic similarity measures across different ontology configurations (MedDRA, MeSH, SNOMED-CT), yielding 40 SSM-ontology combinations. For each SMQ-SSM pair, we applied agglomerative clustering, identifying the similarity threshold required to include all SMQ PTs, then measuring how many external PTs (outside the SMQ) exceeded that threshold. As shown in Figure 1, performance was variable across the different measures.  
				
			\begin{figure}[htbp]
				\centering
				\fbox{\includegraphics[width=0.85\textwidth]{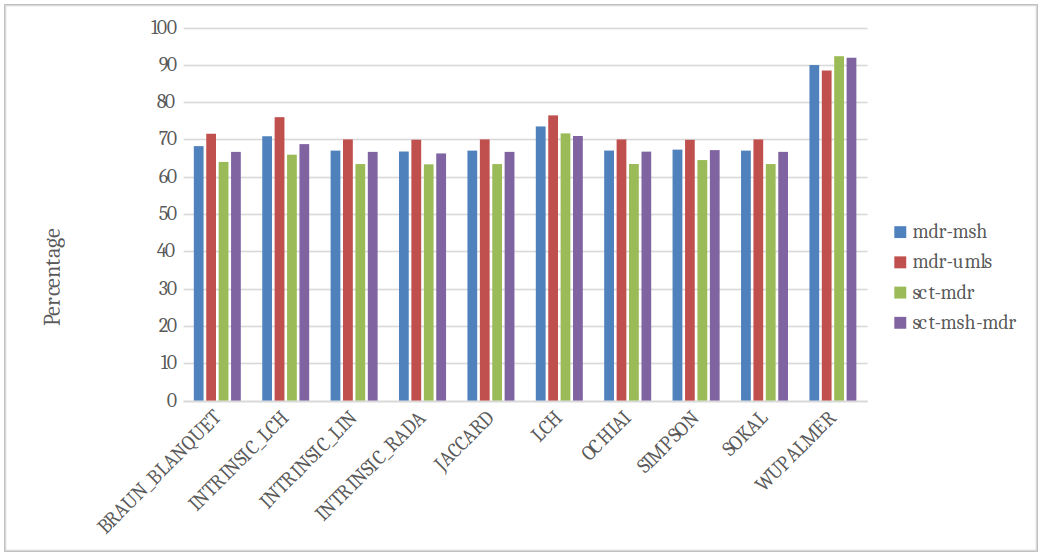}}
				\caption{Feasibility analysis of SSMs and ontological views.  MDR: MedDRA; MSH: MeSH; SCT: SNOMED-CT; UMLS: Unified Medical Language System}
				\label{fig:fig_01}
			\end{figure}
						
			Results indicated that, except for the two path-based metrics (LCH and WUPALMER), the average proportion of external PTs was consistently lower when using MedDRA+SNOMED-CT integrated ontologies. Among the different SSMs, intrinsic IC-based metrics generally included fewer external PTs compared to path-based metrics. However, no clear distinction was observed between the different intrinsic IC-based metrics.  Among path-based methods, LCH and WUPALMER consistently had lower performance, with WUPALMER achieving the weakest performance across all ontology configurations.  
			
			Ultimately, the final six selected measures – LCH, WUPALMER, INTRINSIC\_RESNIK, INTRINSIC\_LIN, INTRINSIC\_LCH, and SOKAL – were chosen based on their strong feasibility study performance, established literature support, and methodological complementarity (path-based vs. IC-based approaches). This selection enables a comparative analysis of path-based and IC-based methods, ensuring a rigorous evaluation of SSMs for pharmacovigilance applications.  
			
			Based on these findings, we refined our evaluation to focus on MedDRA+SNOMED-CT integrated ontologies and the selected six metrics, aligning with previously reported performance trends \cite{pedersen2007measures}.

				\subsection{Medical Review and Reference Standards}
				
				Six medical experts independently reviewed PT lists generated per centroid PT. Each list was assessed by two reviewers, with discrepancies resolved by a senior expert. Reviewers underwent a calibration training session and were asked to judge whether each PT should be retrieved when searching for the associated adverse event concept (e.g., Interstitial lung disease). 
				
				Twelve SMQs were reviewed (Table \ref{tab:smq}), each anchored by a centroid PT matching the SMQ name, with PTs sourced from three categories: SMQ terms explicitly listed under the SMQ, PTs sharing a High-Level Group Term (HGLT) with the centroid, and proprietary GSK-defined groupings referred to as ``autolist'' terms.
				
				These formed the non-semantic distance based inclusion test set. Each SMQ averaged 88 PT codes based on the criteria above (median = 82.5). Many of these additional terms are attributed to the broad scope of HLGT inclusion. SSM-derived PTs – those retrieved only by semantic similarity – were treated as \textit{negative controls}.

				\begin{table}
					\centering
					\caption{SMQs included in the evaluation.}
					\begin{tabular}{|c|c|l|r|r|}
						\hline
						\textbf{SMQ ID} & \textbf{Centroid PT} & \textbf{SMQ Name} & \textbf{Control PTs} & \textbf{Total PTs} \\
						\hline
						20000024 & 10002424 & Angioedema & 101 & 580 \\
						20000004 & 10007554 & Cardiac failure & 64 & 429 \\
						20000150 & 10007636 & Cardiomyopathy & 136 & 542 \\
						20000073 & 10012267 & Dementia & 77 & 520 \\
						20000154 & 10012305 & Demyelination & 77 & 615 \\
						20000146 & 10018304 & Glaucoma & 63 & 450 \\
						20000042 & 10022611 & Interstitial lung disease & 128 & 569 \\
						20000217 & 10028533 & Myelodysplastic syndrome & 107 & 582 \\
						20000047 & 10028596 & Myocardial infarction & 66 & 455 \\
						20000222 & 10038695 & Respiratory failure & 117 & 584 \\
						20000045 & 10042945 & Systemic lupus erythematosus & 55 & 473 \\
						20000238 & 10040477 & Sexual dysfunction & 88 & 548 \\
						\hline
						\multicolumn{3}{|r|}{\textbf{Total}} & \textbf{1,079} & \textbf{6,347} \\
						\hline
					\end{tabular}
					\label{tab:smq}
				\end{table}
							
		\subsection{Constructing the Review Datasets}
\begin{figure}[htbp]
				\centering
				\fbox{\includegraphics[width=0.85\textwidth]{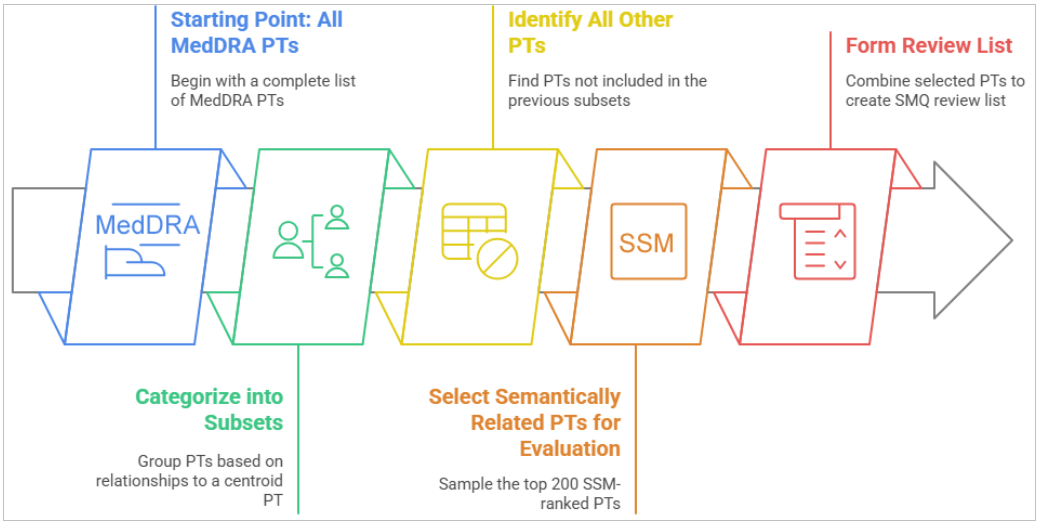}}
				\caption{Process for constructing evaluation dataset}
				\label{fig:fig_02}
			\end{figure}

			For each centroid PT, similarity scores to all MedDRA PTs (n=26,409) were computed and stored in CSV format. To control reviewer burden, we sampled the top 200 SSM-ranked\footnote{SSM measures were computed exclusively under the MedDRA+SNOMED-CT ontological view.} PTs (excluding non-semantically related PT codes) per centroid. If the 200th ranked PT had more than 50 terms tied at the same score, a maximum of 50 were randomly sampled to maintain diversity, while preventing reviewer overload. On average, 529 PTs were evaluated per SMQ (median = 548). The full selection process for the set of review codes is outlined in Figure \ref{fig:fig_02}. 				

		\subsection{Performance Metrics and Review Process}
		
			SSM performance was evaluated against expert judgments using receiver operating characteristic (ROC) curves and precision-recall (P-R) curves. F1-scores were computed across varying similarity thresholds and averaged across all SMQs. To assess how well SSM-based clustering aligned with existing pharmacovigilance (PV) frameworks, we compared clusters against SMQ and HLGT classifications. Cohen’s kappa was used to measure agreement with manual review.
			
		\subsection{Review Process and Adjudication}
		
			To reduce reviewer fatigue, PT lists were randomized and split into two equal groups while maintaining ranking integrity within similarity blocks. Across 12 SMQs, a total of 6,347 PTs were manually reviewed, requiring 218 hours of expert evaluation. Discrepancies were adjudicated, and Cohen’s kappa was computed per centroid PT to benchmark agreement with SMQ and HLGT groupings, establishing a baseline for SSM-based clustering performance.			

	\section{Results}	
			
\begin{figure}[htbp]
				\centering
				\fbox{\includegraphics[width=0.5\textwidth]{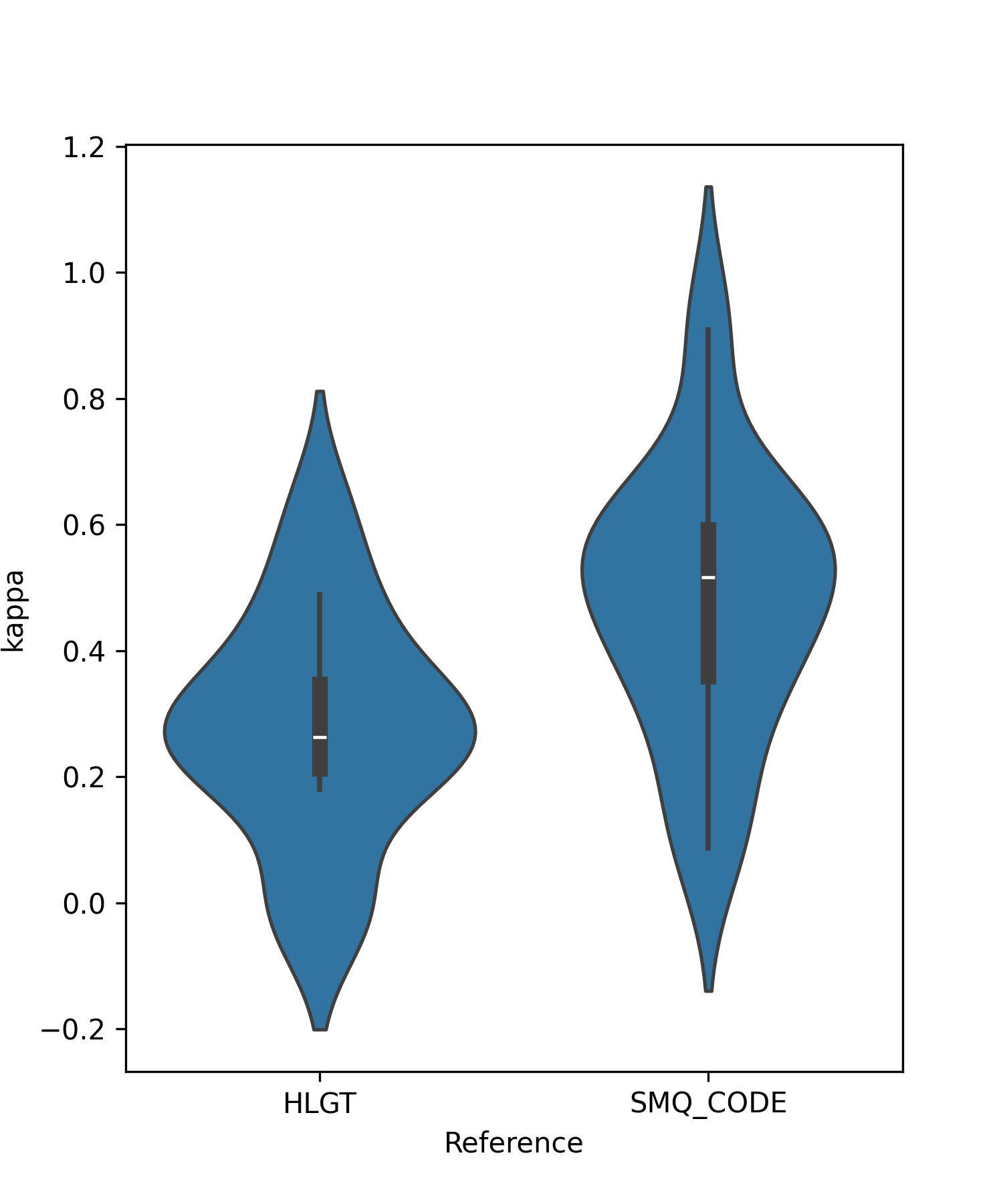}}
				\caption{Cohen’s Kappa comparing medical review to SMQ and HLGT groupings}
				\label{fig:fig_03}
			\end{figure}

		Figure \ref{fig:fig_03} presents the distribution of agreement (Cohen’s Kappa) between medical review and both SMQ and HLGT classifications. Overall, agreement was higher with SMQs, though the variability between centroid PTs was substantial.
			
		Since the evaluated SSMs are not inherently normalized to [0,1], we rescaled them based on their minimum and maximum values across all PTs. The resulting distributions (Figure 4(a)) revealed bimodal patterns for INTRINSIC\_RESNIK, INTRINSIC\_LIN, INTRINSIC\_LCH, and SOKAL. Initially, we hypothesized that the integration of MedDRA and SNOMED-CT contributed to this effect; however, a similar multi-modal distribution was observed when using SNOMED-CT alone (Figure 4(b)), suggesting a more complex underlying structure.

			\begin{figure}[htbp]
				\centering
				\fbox{\includegraphics[width=0.9\textwidth]{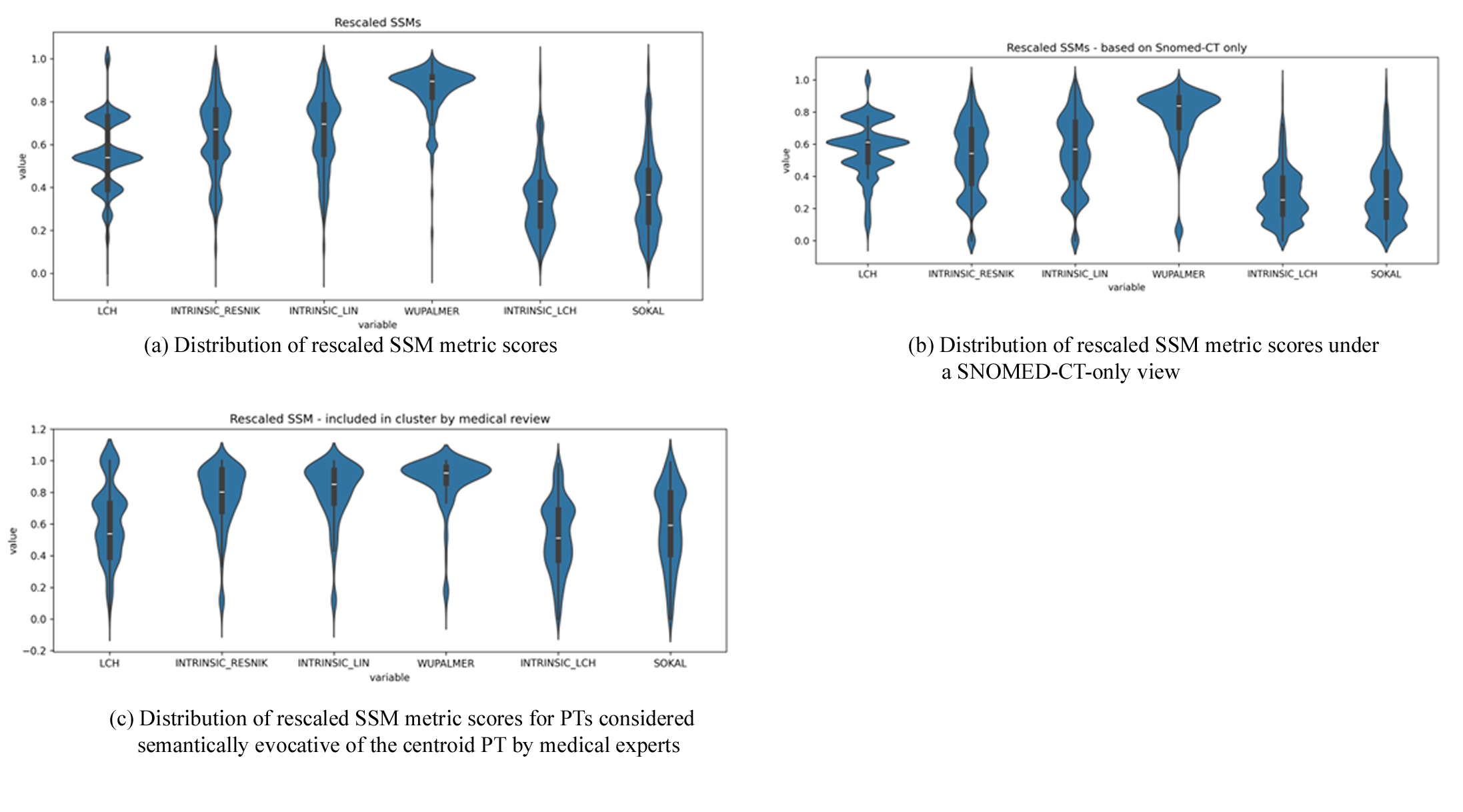}}
				\caption{SSM Distributions}
				\label{fig:fig_04}
			\end{figure}

		PTs included in the expert-determined clusters had higher SSM values than excluded PTs (Figure 4(c)). The difference in distributions was more pronounced for INTRINSIC\_LCH and SOKAL, likely due to their narrower and lower-value distributions.
		
		Figure 5(a) displays the mean ROC curves across centroid PTs based on medical review. Path-based SSMs (LCH, WUPALMER) underperformed compared to IC-based SSMs. Notably, LCH exhibited a step-like ROC curve due to its limited set of distinct values. Among IC-based SSMs, INTRINSIC\_LIN and SOKAL produced identical curves due to their strictly monotonic relationship. INTRINSIC\_RESNIK performed less favorably than the other IC-based measures.
		
		When SMQs were used as the reference standard ROC curves (Figure 5(b)) exhibited similar performance for all IC-based SSMs. However, past a false positive rate of 0.75, these curves dipped below the identity line. Under this criterion, LCH showed closer alignment with IC-based curves, whereas WUPALMER’s performance deteriorated more sharply. HLGT-based results (Figure 5(c)) closely resembled those obtained using medical review.
		
		The mean precision-recall (P-R) curves derived from medical review revealed that the positive predictive value (PPV) of INTRINSIC\_LIN, INTRINSIC\_LCH, and SOKAL was lower at sensitivity levels below 0.5 but subsequently increased (Figure 5(d)). In contrast, WUPALMER exhibited the opposite trend. This pattern became clearer when examining the F1-score as a function of SSM values (Figure 6). The highest F1-score was achieved by INTRINSIC\_LCH (0.404), followed closely by INTRINSIC\_LIN and SOKAL, both scoring 0.403. The path-based SSMs achieved lower performance with F1 scores of 0.36 for WUPALMER and 0.28 for LCH. Notably, the F1-score curve of INTRINSIC\_LIN mirrored that of SOKAL but was shifted toward higher SSM thresholds. 
		
		When SSM thresholds increased, the increase in F1 score seemed to follow mainly a two steps shape, with a first sharp increase between from 0.2 to 0.3 for INTRINSIC\_LCH and SOKAL and between 0.55 and 0.7 for INTRINSIC\_RESNIK and INTRINSIC\_LIN. The second sharp increase was between 0.4 and 0.45 for INTRINSIC\_LCH, between 0.45 and 0.5 for SOKAL, and between 0.7 and 0.8 for INTRINSIC\_RESNIK and INTRINSIC\_LIN.		
		
			\begin{figure}[htbp]
				\centering
				\fbox{\includegraphics[width=0.95\textwidth]{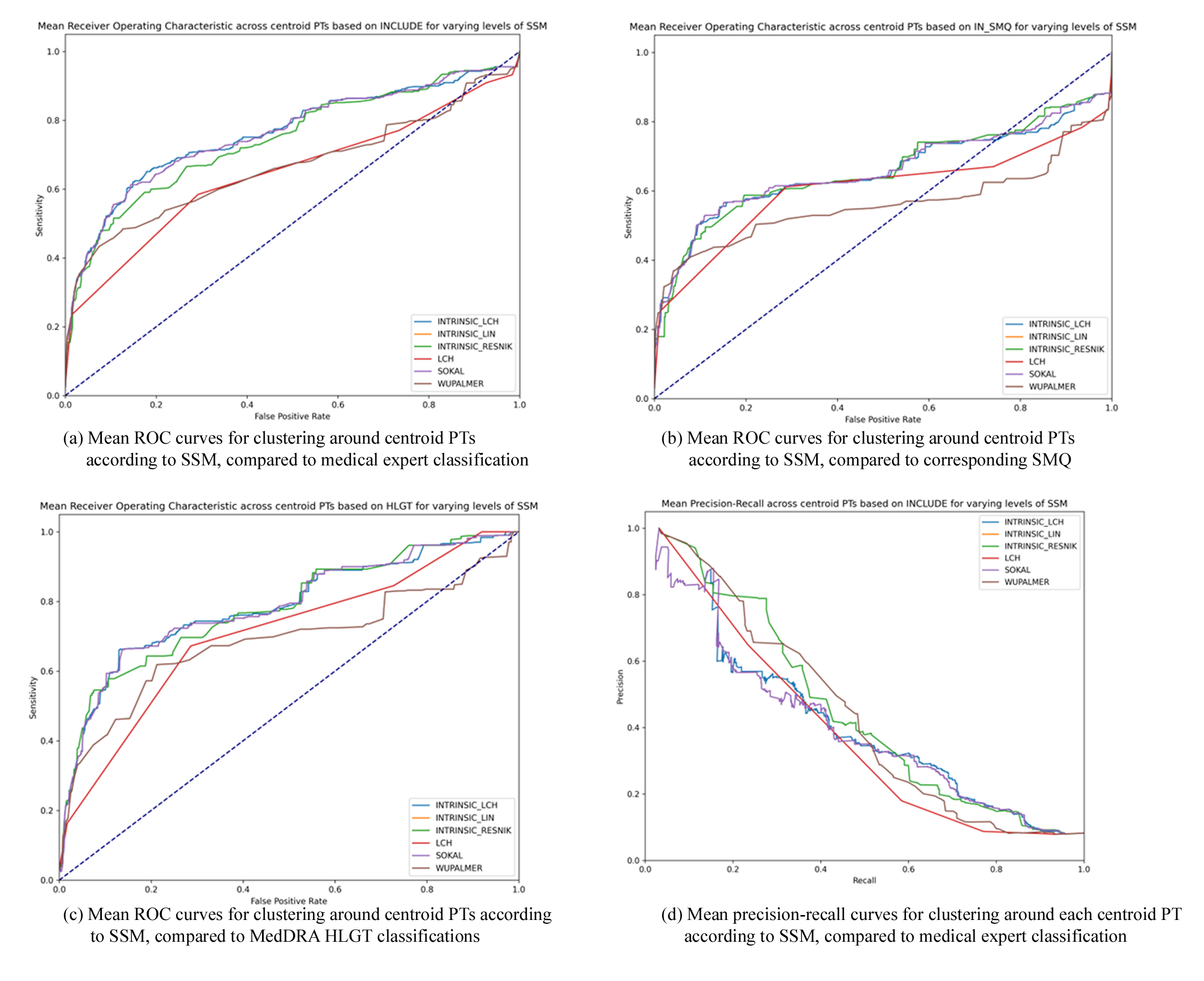}}
				\caption{ROC Curve Analyses}
				\label{fig:fig_05}
			\end{figure}

		F1-scores computed using SMQ-based evaluation were generally lower than those based on medical review, with a maximum score of 0.367 for the three best-performing SSMs. Similarly, HLGT-based evaluation yielded slightly lower F1-scores, with 0.398 for INTRINSIC\_LCH and 0.387 for both INTRINSIC\_LIN and SOKAL.

			\begin{figure}[htbp]
				\centering
				\fbox{\includegraphics[width=0.95\textwidth]{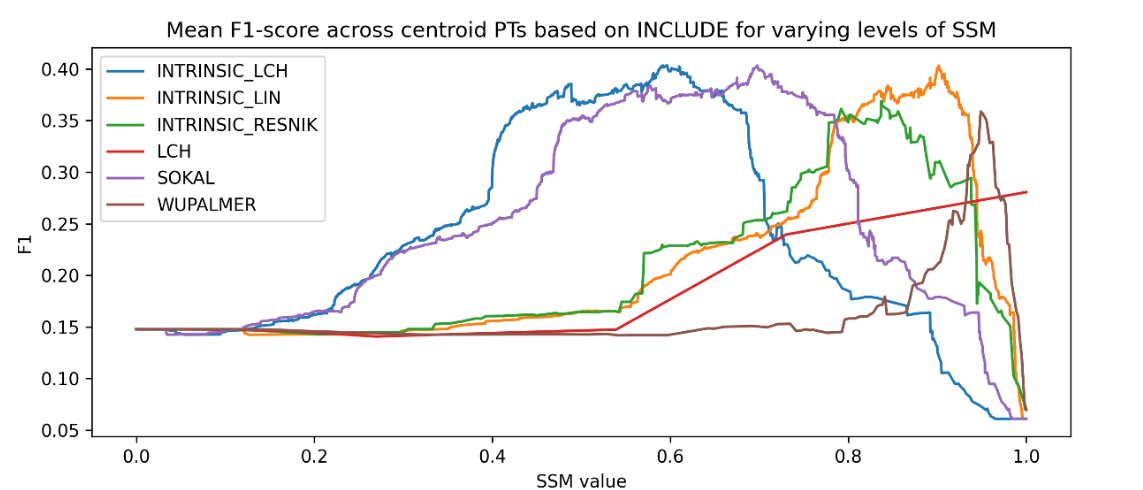}}
				\caption{F1-score as a function of SSM thresholds for clustering around centroid PTs according to each SSM, compared to medical expert classification}
				\label{fig:fig_06}
			\end{figure}

	\section{Discussion}	
			
		Our evaluation demonstrates that semantic similarity measures (SSMs), when carefully applied to drug safety data, can effectively cluster Preferred Terms (PTs) around clinically relevant centroids. Among the six tested SSMs, intrinsic information content (IC)-based measures—particularly INTRINSIC\_LCH, INTRINSIC\_LIN, and SOKAL—consistently outperformed path-based approaches. These measures likely benefit from incorporating both IC and conceptual overlap, enabling more nuanced differentiation between related terms. In contrast, the relatively lower performance of INTRINSIC\_RESNIK suggests that using only the IC of the lowest common ancestor (LCA) may be insufficient for fine-grained clustering.
		
		Interestingly, several SSMs exhibited bimodal or multimodal similarity score distributions, even within SNOMED-CT. These patterns likely reflect the influence of ontology-specific properties, such as term granularity and concept density. Future work may explore ontology preprocessing techniques, including branch pruning or IC recalibration, to address such distributional artifacts.
		While SSM-based clustering approximated expert judgment, it did not fully replicate it. Comparisons with established references (SMQs, HLGT) revealed that HLGT-based groupings unexpectedly outperformed SMQs in some cases, potentially due to the composition of the selected SMQs. These findings reinforce the value of SSMs as decision-support tools that complement—but do not replace—clinical expertise.
		
		Our evaluation framework—anchored by expert-reviewed centroid PTs—ensures practical relevance but introduces subjectivity, echoing challenges identified by Pedersen et al. \cite{pedersen2007measures} regarding alignment between expert raters. Moreover, the number of centroids evaluated was constrained by the availability of medical reviewers, a common limitation in semantic similarity studies.
		
		An important limitation of this study—and of most PT-level clustering analysis is the assumption that MedDRA PTs consistently and accurately reflect the underlying clinical events. In practice, the selection of PTs in safety reports can be influenced by reporter expertise, site conventions, or data entry constraints, introducing variability that may affect both clustering outcomes and evaluation. As a result, even perfectly grouped PTs may not always correspond to distinct or consistent clinical phenomena. Addressing this limitation may require incorporating contextual data from narratives or leveraging probabilistic models to account for PT variability in future work.
		
	\section{Conclusion}

		This study presents a systematic evaluation of six ontology-based semantic similarity measures (SSMs) for clustering MedDRA PTs in a PV setting. IC-based metrics, especially INTRINSIC\_LIN, INTRINSIC\_LCH, and SOKAL, demonstrated strong potential for improving automated clustering of MedDRA PTs. These methods offer a practical path toward reducing manual curation workloads and enhancing early signal detection in large-scale drug safety analyses.
		
		Future research may expand the ontology base beyond MedDRA and SNOMED-CT, explore hybrid models that integrate lexical or contextual similarity, or combine SSMs with statistical signal detection metrics to improve overall predictive accuracy

	\section{Declarations}
	
		GSK covered all costs associated with the conduct of the study and the development of the manuscript and the decision to publish the manuscript. J.P., F.H., G.P, and A.B. are employed by GSK and hold financial equities. This manuscript has not been submitted to, nor is under review at, another journal or other publishing venue. The authors have no competing interests to declare that are relevant to the content of this article.

		\textbf{Data availability} All data and code to support this analysis is available from \url{https://github.com/jlpainter/AMIA2025/tree/main/ssm}.\\		
	
		\section{Acknowledgments}
			
			We acknowledge the UNC Eshelman School of Pharmacy PharmD students for reviewing SSM data under supervision of Greg Powell, PharmD: Miranda Barker, Allen Barry, Brady Carlton, Megan Earnhart, Matthieu Soares, and Caroline Todd.

	\bibliography{ssm_metric}

\end{document}